\documentclass[11pt]{article}

\usepackage[final]{acl}

\usepackage{times}
\usepackage{latexsym}
\usepackage{enumitem}

\usepackage[T1]{fontenc}
\usepackage[utf8]{inputenc}

\usepackage{microtype}
\usepackage{inconsolata}
\usepackage{graphicx}
\usepackage{booktabs} 
\usepackage{float}    
\usepackage{multirow}
\usepackage{tabularx}
\usepackage{nicematrix}
\usepackage[most]{tcolorbox}

\usepackage{amsmath}
\usepackage{amssymb}
\usepackage{mathtools}

\usepackage{algorithm}
\usepackage{algorithmic}

\title{Streaming Translation and Transcription \\ Through Speech-to-Text Causal Alignment}

\author{
  \textbf{Roman Koshkin, Jeon Haesung, Lianbo Liu, Hao Shi,} \\
  \textbf{Mengjie Zhao, Yusuke Fujita, Yui Sudo} \\
  SB Intuitions, Tokyo, Japan \\
  \small \textbf{Correspondence:} \href{mailto:roman.koshkin@sbintuitions.co.jp}{roman.koshkin@sbintuitions.co.jp}
}

\begin{document}
\maketitle

\begin{abstract}
Simultaneous machine translation (SiMT) has traditionally relied on offline machine translation models coupled with human-engineered heuristics or learned policies. We propose Hikari, a policy-free, end-to-end model for simultaneous speech-to-text translation and streaming transcription. We also introduce Decoder Time Dilation, a mechanism that counteracts the overrepresentation of WAIT tokens in training. We present a supervised fine-tuning strategy that trains the model to recover from delays, significantly improving the quality-latency trade-off.  Despite its modest size, Hikari delivers competitive translation quality at consistently low latency, comparing favorably with published IWSLT 2026 submissions up to ~38× larger and with proprietary API systems across en→ja, en→de, and en→ru. We release our model weights and code to facilitate further research.
\end{abstract}

\section{Introduction}
\label{introduction}

Simultaneous translation serves audiences who need speech rendered in another language in near real time. A good simultaneous translation must not only be accurate and complete, but also not lag unacceptably far behind the original message. This requires interpreters to strategically decide whether to translate what has been said so far  -- often making assumptions about the speaker's intended message -- or delay translation until more information is revealed by later context \citep{Ilyukhin2001, chernov2004inference, setton2005pointing, AMOS2022104987}.

\begin{figure}[ht!]

\vskip 0.2in
   \centering   \includegraphics[width=0.99\columnwidth] {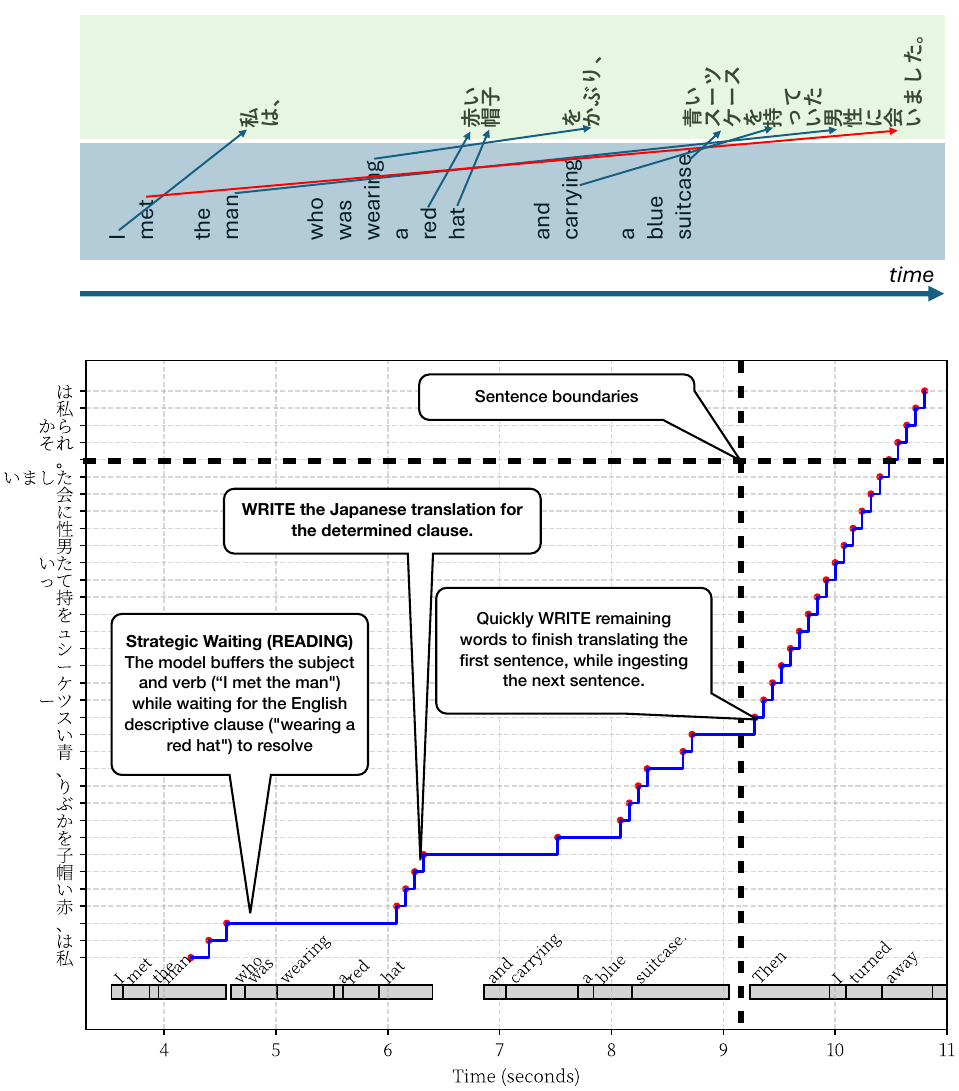}

   \caption{\label{fig:hero} The challenge and reality of simultaneous translation. (Top) Conceptual illustration of syntactic divergence between English (SVO) and Japanese (SOV); the translator must hold the main verb "met" until the end of the Japanese sentence. (Bottom) Actual inference timeline, showing how Hikari has learned the strategic patience required for syntactically divergent languages. Horizontal segments represent the model waiting for context (READING, i.e. emitting \texttt{WAIT} tokens not shown for clarity), while the steep "steps" are rapid token emission (WRITING) once semantic ambiguity is resolved.}
\vskip -0.1in
\end{figure}

This decision-making is especially challenging for syntactically divergent language pairs such as English and Japanese. English provides the principal subject and verb early (SVO), whereas Japanese, a head-final (SOV) language, requires these elements -- along with complex relative clauses -- to be restructured. As illustrated in Figure \ref{fig:hero} (top), when translating the sentence "I met the man who was wearing a red hat and carrying a blue suitcase," the translator (human or machine) must remember/buffer the core components (the subject and the main verb "met") for several seconds while it processes the long English descriptive clause. The Japanese output cannot be safely committed until the entire qualifying phrase ("red hat wearing, blue suitcase carrying") has been heard, resulting in an inherent temporal lag.

Prior work models this decision-making with either learned READ/WRITE\footnote{where "READ" and "WRITE" refer to ingesting more input audio and committing partial translation, respectively.} policies or human-crafted heuristics. These approaches segment input audio (or text, in cascaded systems) along semantic or syntactic boundaries and pass each chunk to a pre-trained MT model.

Hikari is instead policy-free: READ and WRITE actions are encoded in the next-token distribution. If the model is uncertain, the probability of a text token drops, and the probability of a special \texttt{WAIT} token rises. This unifies the "what" (translation) and the "when" (policy) into a single probabilistic framework. Thus, Hikari jointly learns an audio-conditioned language model and an implicit READ/WRITE policy. Our contributions are summarized as follows:

\begin{itemize}[itemsep=-0.45em, topsep=0pt]
  
	\item We propose a unified speech-to-text framework that jointly learns translation and transcription without external READ/WRITE policies.

	\item We introduce Decoder Time Dilation, a mechanism that addresses the dominance of \texttt{WAIT} tokens in training data by increasing the decoder's effective step size relative to the audio.

	\item We develop a fine-tuning method that trains the model to actively minimize lag and recover from latency drift during inference.

	\item Despite its modest size (0.76 B parameters — up to ~38× smaller than the compared IWSLT systems), Hikari attains competitive translation quality at the lowest latency among all systems on en→de and en→ru, and competitive latency on en→ja.

	\item We provide open access to our model weights and inference code.

\end{itemize}

\section{Related Work}

\subsection{Cascaded Systems and Heuristic Policies}

Traditionally, simultaneous speech-to-text translation (SiMT) has relied on cascaded architectures where an offline\footnote{Offline MT models translate after processing the full input, rather than incrementally.} machine translation (MT) model (e.g. mBART \citep{liu-etal-2020-multilingual-denoising}, originally developed for sentence-level translation) is paired with human-engineered heuristics. Early systems used fixed policies such as "wait-k" \citep{ma-etal-2019-stacl} or "wait-if" \citep{cho2016neural}. However, these policies struggle, especially with syntactically divergent language pairs. If a critical verb appears at the end of a German sentence (SOV structure in subordinate clauses), a \textit{wait-k} model with a small $k$ is forced to hallucinate the verb to satisfy the policy, resulting in catastrophic quality loss. Conversely, a large $k$ incurs unacceptable latency.

\subsection{Adaptive and Learned Policies}

To address the rigidity of heuristics, subsequent research introduced learned policies \citep{alinejad-etal-2018-prediction, Satija2016SimultaneousMT, grissom-ii-etal-2014-dont, gu-etal-2017-learning}, attention-based methods \citep{Ma2020Monotonic, Papi2023, papi-etal-2023-attention, arivazhagan-etal-2019-monotonic, 10.5555/3305890.3305974, DBLP:conf/iclr/ChiuR18}, binary search strategies \citep{guo-etal-2023-learning}, and approaches leveraging the stability of output predictions through "local agreement" \citep{polak-etal-2022-cuni, ko-etal-2023-tagged, liu20s_interspeech}. While more adaptive, these methods either introduce architectural complexities or involve training a controller disconnected from the translation model itself.

\begin{figure*}[ht!]
\vskip 0.2in
   \centering
   \includegraphics[width=0.85\textwidth] {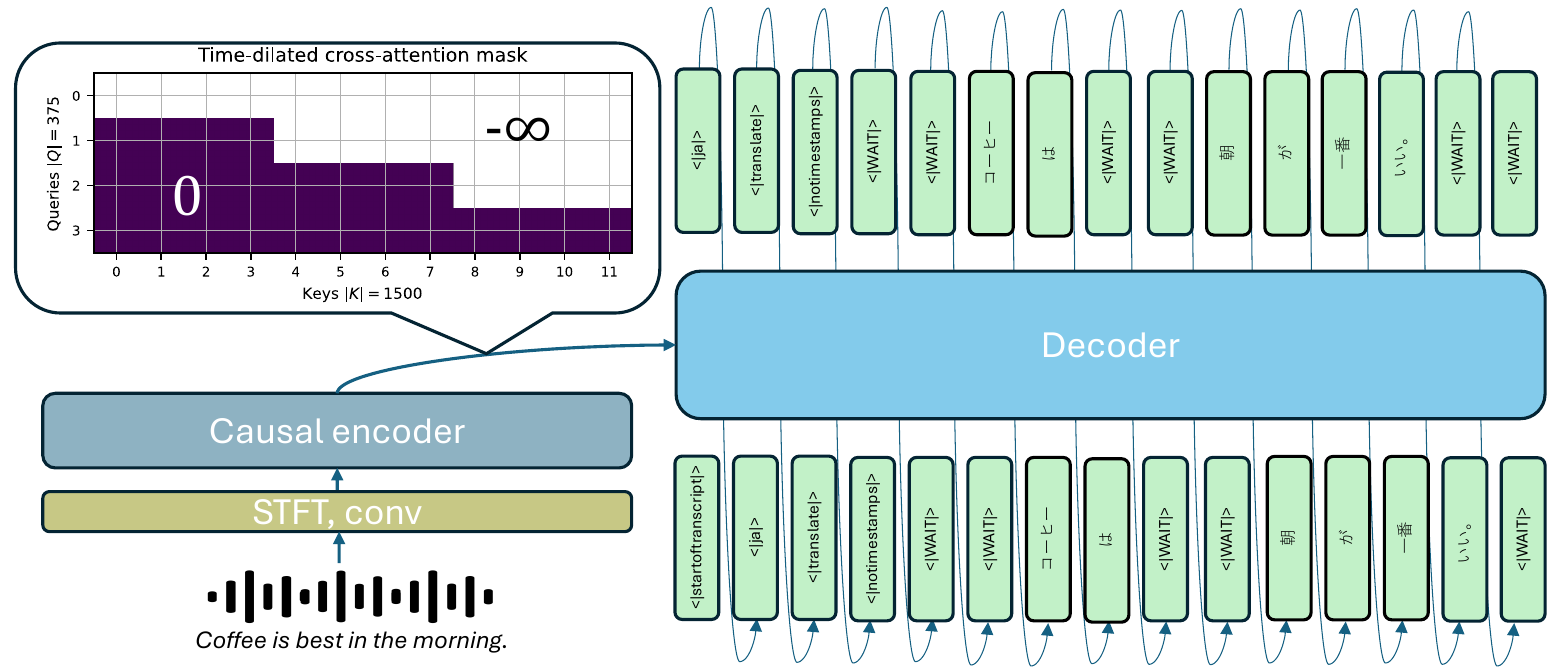}
   \caption{\label{fig:architecture} High-level overview of the model architecture.}
\vskip -0.2in
\end{figure*}

\subsection{Foundation Models and SiMT}

Recent SiMT systems have increasingly incorporated large-scale foundation models -- such as Qwen \citep{qwen2025qwen25technicalreport}, Gemma 3 \citep{gemmateam2025gemma3technicalreport}, Llama 2 \citep{touvron2023llama2openfoundation}, and NLLB \citep{nllb2024scaling} -- into both cascaded and end-to-end frameworks. Systems like BeaverTalk \citep{raffel-etal-2025-beavertalk} employ conversational prompting with memory banks, while InfiniSST \citep{ouyang-etal-2025-infinisst} and the NAIST 2025 system \citep{tan-etal-2025-naist} use chunk-wise causal encoders or external segmentation tactics such as SHAS \citep{tsiamas22interspeech}. Despite their scale, these systems often still rely on modular components -- e.g., a separate VAD model \citep{machacek-polak-2025-simultaneous,shan-etal-2024-hw} -- to govern when to emit text, which can complicate inference on unbounded speech streams.

This trend has been confirmed by the most recent IWSLT evaluation campaign \citep{adelani-etal-2026-speech}, which reports a clear shift toward LLMs across submitted systems, with cascaded architectures dominating and only a single end-to-end submission. Notably, higher latency in large-model systems does not always yield proportional translation quality gains. This points to an underexplored opportunity: small, fully end-to-end SiMT models such as Hikari -- trained with explicit READ/WRITE supervision via causal alignment -- offer a competitive and low latency alternative.

\subsection{Policy-Free Paradigm}

In a notable departure from modular designs, \citet{koshkin2024transllama} proposed \textit{causal alignment}, a data construction method that encodes READ/WRITE supervision directly in the training examples. Their cascaded SiMT system fine-tuned an LLM to perform translation and determine when to READ or WRITE within a single model. More recent systems have adopted a similar principle, learning streaming behavior from temporally aligned training data rather than relying on an external policy \citep{labiausse2025highfidelity,labiausse2026simultaneousspeechtospeechtranslationaligned}.

Hikari extends causal alignment to the speech-to-text domain, yielding a policy-free, fully end-to-end SiMT architecture. As illustrated in Figure \ref{fig:hero}, the model strategically pauses for nearly two seconds while ingesting a complex English relative clause, only to "burst" the corresponding Japanese translation the moment the head noun ("hat") provides semantic closure. Hikari also introduces \emph{Decoder Time Dilation} to solve the overrepresentation of filler tokens -- a known problem in alignment-based streaming systems \citep{labiausse2025highfidelity}.

\section{Method}

\subsection{Architecture}

Our architecture (Figure \ref{fig:architecture}) is based on the Whisper encoder-decoder Transformer \citep{pmlr-v202-radford23a}. While the original Whisper model is designed for \emph{offline} processing -- requiring the entire 30 s audio window to be encoded \emph{before} generation -- we introduce three modifications to enable fully streaming SiMT. First, we encapsulate the READ/WRITE decision-making process into the model's vocabulary by introducing a special \texttt{WAIT} token. In this framework, the emission of \texttt{WAIT} represents a READ action, while the emission of any text token represents a WRITE action. The decoder therefore generates a sequence of READ/WRITE actions that are \emph{temporally synchronized} with the encoder. Second, we make the encoder causal, applying a lower-triangular mask to its self-attention so that each audio frame attends only to earlier frames; this is what allows incremental audio ingestion rather than requiring the full 30 s window up front. Third, to prevent the decoder from attending to the encoder's future keys during training, we apply a \emph{dilated} causal mask to the cross-attention.

\textbf{Decoder time dilation.} As in the original Whisper model, the input waveform (mono, sampled at 16000 Hz, padded to 30 s) is transformed into a log-mel spectrogram with 80-dimensional feature vectors, each capturing 10 ms of input audio. The spectrogram is temporally downsampled by a factor of 2 by the convolutional layers to 1500 audio embeddings which are fed into the encoder.

\begin{figure}[ht!]
\vskip 0.1in
\centering
\includegraphics[width=0.90\columnwidth] {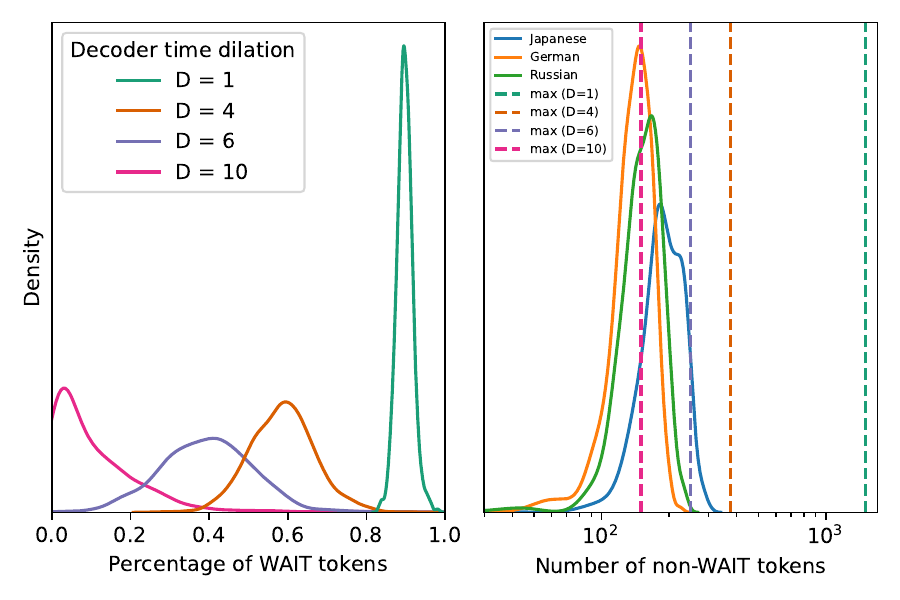}
\caption{\label{fig:fig_hist} Choosing the optimal value of decoder time dilation ($D$). Left panel:  percentage of wait tokens in a 30 s en-ja sample as a function of $D \in \{1,4,6,10\}$. Lower $D$ improves precision of causal alignment, but also (undesirably) increases the dominance of \texttt{WAIT} tokens in the training data. Right panel: with too high a value of $D$ (e.g. 6 and 10), target tokens corresponding to 30 s of input audio might not fully fit into the decoder's context (shown by dashed vertical lines for $D \in \{1,4,6,10\}$.}
\vskip -0.1 in
\end{figure}

While the original Whisper model generates output tokens autoregressively based on the \emph{entire} 30 s window, our model should generate a new token for each new $D$ audio embeddings. We could simply set $D=1$, but in that case about 90\% (Figure \ref{fig:fig_hist}) of the tokens that the decoder is expected to generate would be \texttt{WAIT} tokens\footnote{Since \texttt{WAIT} doesn't exist in Whisper's tokenizer, we use token 93, corresponding to the tilde character ("$\sim$") that never occurs in our training set, as \texttt{WAIT}.}, often appearing in long sequences uninterrupted by non-wait tokens. We found such an imbalance (with $D < 4$) to cause the model to excessively generate those \texttt{WAIT} tokens at inference. One way to address this is either to penalize the loss on the \texttt{WAIT} token at training or reduce the log-likelihood of the corresponding token at inference. We found empirically, that it is better to instead "dilate" decoder time by choosing $D > 1$. This reduces the output token rate, the frequency of \texttt{WAIT}, as well as the number of forward passes through the decoder corresponding to a 30 s of input audio. Valid choices of $D$ range between 1 and approximately 4 (Figure \ref{fig:fig_hist}). Setting a larger value of $D$ would increase the length of audio corresponding to one decoder token (e.g. with $D=10$ one output token would correspond to 200 ms of speech), not only reducing the temporal precision of causal alignment, but also (and most importantly), making it impossible for all the target tokens to always fit into the decoder's context window at high rates of speech (in WPM). With $D=4$, all the tokens always fully fit in the decoder's context (Figure \ref{fig:fig_hist}), ensuring a good tradeoff between, on the one hand, the granularity of causal alignment, and proportion of \texttt{WAIT} tokens in the training set and the whole model's real-time factor (RTF) (see Appendix \ref{appendix:rtf},  Figure \ref{fig:RTF}), on the other. Furthermore, we analyze the model's throughput under batched conditions (Figure \ref{fig:RTF}), which is critical for scalable deployment. Hikari requires $D \ge 4$ to support batch sizes up to 10 on a single H100 GPU in real time. At $D < 4$, the computational cost of the autoregressive decoder makes high-concurrency serving impractical (RTF $> 1.0$), and therefore, we report our results obtained with $D=4$. Thus, the cross-attention mask is as follows:

$$
M_{m,n} = \begin{cases*}
0 & if $n \leq 4(m-1)$ \\
-\infty & otherwise
\end{cases*}
$$

where the decoder's queries and encoder's keys are indexed by $m \in \{1, \dots, 375\}$ and $n \in \{1, \dots, 1500\}$, respectively. This fixed dilation factor resembles the fixed-ratio synchronization used in MinMo \citep{chen2025minmomultimodallargelanguage}.

\subsection{Data}

Ideally, Hikari would be trained on authentic audio recordings paired with professional simultaneous interpretations. However, such corpora are rare and mostly limited to bureaucratic and diplomatic discourse in official UN languages\footnote{Arabic, Chinese, English, French, Russian, and Spanish}, which might not generalize to spontaneous speech or other domains. Instead, we developed a pipeline to generate synthetic training data, establishing temporal correspondences between source audio and target translations through the causal alignment process detailed in Section \ref{subsec:causal_alignment}.

\subsubsection{Preparation}
Briefly, the data preparation pipeline involves the following three steps. First, we use Faster Whisper\footnote{https://github.com/SYSTRAN/faster-whisper} to obtain transcripts with word-level timestamps of the input audio, such that each word and its immediately following punctuation character (if any) has its own start and end timestamp. Second, we split the transcript into sentences, and translate them one by one with an LLM. To prevent discourse-level inconsistencies (e.g.\ in terminology or pronoun reference), the LLM is supplied with a rolling context of four preceding sentences (see Appendix \ref{appendix_A} for the full prompt template). Third, for each source and its corresponding target sentence, we perform the causal alignment procedure \citep{koshkin2024transllama} which consists in establishing correspondences between the source and target words\footnote{For Japanese targets we use the \texttt{janome} package to split the translation into word-like character strings.} based on the similarity of source and target word embeddings \citep{jalili-sabet-etal-2020-simalign}.

\subsubsection{Causal alignment}
\label{subsec:causal_alignment}

Let the transcript of one source sentence be represented as a sequence of $M$ words $X = (x_1, x_2, \dots, x_M)$, where each word $x_i$ is associated with a time interval $[s_i, e_i]$ obtained via word-level timestamps. Here, $s_i$ and $e_i$ denote the start and end times of the word in milliseconds, respectively. Using a word embedding-based alignment model \citep{jalili-sabet-etal-2020-simalign}, we establish a correspondence mapping $f: \{1, \dots, N\} \to \{1, \dots, M\} \cup \{\emptyset\}$ between the target words $Y = (y_1, y_2, \dots, y_N)$ and the source words $X$. For any target word $y_j$, $f(j) = i$ indicates that $y_j$ is the translation of source word $x_i$. To ensure the model is truly simultaneous, we define a target release time $t(y_j)$ for each target word. The source and target words are considered causally aligned if:

$$t(y_j) \ge e_{f(j)} + \delta$$

where $e_{f(j)}$ is the end timestamp of the corresponding source word $x_i$ and $\delta$ is a stochastic delay $\delta \sim \mathcal{U}(0, 200)$ ms, sampled independently for each word to help reduce overfitting. The target token sequence $L$ is constructed by interleaving the original tokenized target words $y_j$ with a special \texttt{WAIT} token. The number of \texttt{WAIT} tokens inserted before $y_j$ is the minimum required such that the "logical time" of each output token matches the physical arrival of the audio information. Target words lacking a direct alignment ($f(j) = \emptyset$) are temporally shifted to occur immediately after the last aligned token. This deferral preserves original word order while ensuring causality is never violated. The formal logic for this synchronization is detailed in Algorithm \ref{alg:causal_alignment} (see Appendix \ref{appendix_D}). Note that for streaming ASR no causal alignment and translation are needed, and the target tokens represent the tokenized source transcript with \texttt{WAIT} tokens inserted according to the corresponding words' timestamps.

\subsubsection{Sources}

Our training set comprises 62.3K hours of public and in-house data (see Table~\ref{tab:datasets-single} in Appendix~\ref{appendix:eval_data} for details). While our primary focus was streaming translation of long-form audio, we still included CommonVoice 2 (CV2) \citep{ardila2020commonvoicemassivelymultilingualspeech} and Multilingual LibriSpeech (MLS) \citep{Pratap_2020}  -- short-form datasets that offer diverse accents, speaking styles, and noise conditions, enhancing the model's robustness. Each audio is translated into German, Japanese and Russian. Thus, we train the model for 4 tasks simultaneously  -- 3 translation tasks (en-de, en-ja and en-ru) and 1 transcription (ASR) -- which are sampled uniformly during training. During training, if the sampled audio file is longer than 30 s (as is the case for podcasts), the dataloader returns a random 30-s section of it with the tokenized target text corresponding to that chunk. If the audio source is shorter than 30 s, it is padded on the right with \texttt{WAIT} to fill the decoder's context window. If the sampled audio section contains fewer than 4 words (e.g. music, silence or noise), the dataloader samples another until there are 4 or more words. The tokenized text is prepended with 4 special tokens: \texttt{<|startoftranscript|>}, target language marker (either \texttt{<|en|>}, \texttt{<|ja|>}, \texttt{<|ru|>} or \texttt{<|de|>}), task marker (either \texttt{<|transcribe|>} or \texttt{<|translate|>} and finally the \texttt{<|notimestamps|>} special token. Thus, assuming $D=4$, to generate the first token, the decoder attends to the past 12 audio embeddings (240 ms of audio, see the cross-attention mask in Figure \ref{fig:architecture}).

\subsection{Training}
\label{sec:training}

\subsubsection{Pre-training}

We initialize both the encoder and decoder from \texttt{whisper-medium}\footnote{https://huggingface.co/openai/whisper-medium}. Although not strictly necessary, our experiments show that such initialization greatly accelerates convergence despite the encoder's self-attention and cross-attention becoming causal. We pre-train the model with AdamW \citep{loshchilov2018decoupled} on 96 H100 GPUs with a per-device batch size of 128 (which approximately corresponds to 102 h of audio) for a total of 35K steps, out of which 100 steps are a linear warmup to a peak learning rate of $0.00004$ followed by a linear decay to zero. We also use gradient accumulation over 2 steps and set the gradient clipping hyperparameter to 1.0.

\subsubsection{Supervised fine-tuning}
\label{sec:sft}
\begin{figure}[h!]
    \vskip -0.16in
   \centering
   \includegraphics[width=0.95\columnwidth] {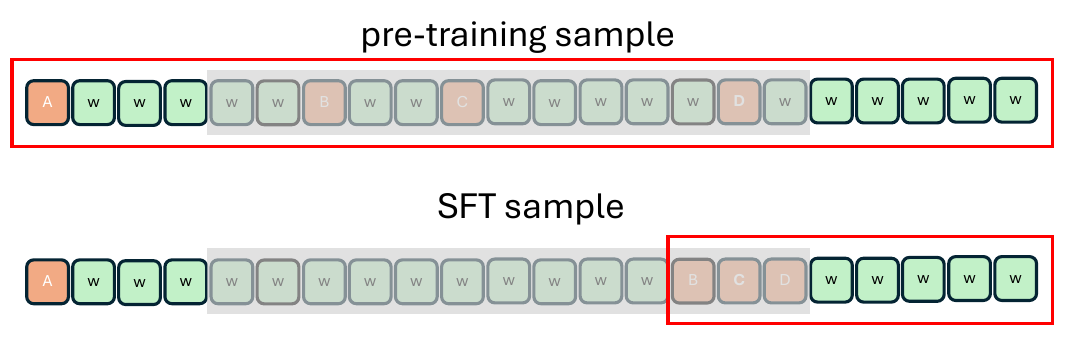}
   \caption{\label{figure3} Construction of an SFT sample. The shaded transparent box indicates the chosen tokens to be delayed. The red rectangles illustrate the tokens for which cross-entropy is calculated. Orange and green squares are non-wait and \texttt{WAIT} tokens, respectively.}
   \vskip -0.1in
\end{figure}

While the model is functional immediately after pre-training, it tends to generate consecutive \texttt{WAIT} tokens, which not only unnecessarily increases translation latency but can also push the decoder state out of distribution, leading to generation failures. To mitigate this, we fine-tune the model on specially constructed samples that simulate recovery from high-latency scenarios. The delay is introduced via a two-step process. First, a pivot index $s_{start}$ is sampled from $\mathcal{U}(4, N)$, where $N$ is decoder's context length. Second, letting $W_{right}$ be the count of \texttt{WAIT} tokens in the segment $y[s_{start}:]$, a displacement $\Delta s \sim \mathcal{U}(1, W_{right}-1)$ is selected. Content tokens following the pivot are deferred by $\Delta s$ positions, effectively accumulating \texttt{WAIT} tokens to force the model to learn rapid "burst" recovery. Figure \ref{figure3} illustrates this process, showing how tokens A, B, C, and D are delayed in a sample sequence. Crucially, we mask the loss for the artificially inserted \texttt{WAIT} tokens and all preceding history. This ensures the model does not learn to imitate this delayed behavior but is instead optimized to catch up. This phase mixes original pre-training samples (70\%) with SFT samples (30\%) to maintain general translation capability while reducing latency.

After the pre-training stage, we fine-tune the model for 2000 steps with the same per-device batch size and warm-up schedule as in pre-training, followed by a constant learning rate of 0.00004.

\subsection{Inference}

To enable translation of long-form audio of unlimited length, we use a fixed-size sliding window at inference. The audio encoder processes input in 80 ms ($20 \text{ ms} \times D$) increments. At each step, a new audio chunk is appended to the buffer; if the total duration is less than 30 s, the input is padded to 30 s. Once the buffer exceeds 30 s, we maintain a sliding window by discarding the oldest audio chunk before appending the new one, ensuring the encoder always receives exactly 30 s of context.

A similar sliding window is applied to the decoder. We use greedy decoding. Generation begins with 4 prompt tokens. As new tokens are generated, they are appended to the context until the sequence reaches a maximum length of 375 tokens. Beyond this limit, for every new token generated, the earliest non-prompt token (i.e., the fifth token in the sequence) is removed. This preserves the initial prompt while keeping the decoder's context window bounded.

\section{Results}

\subsection{Translation quality and latency}
\label{sec:transl_quality_and_latency}

We evaluate against eight systems from IWSLT 2026 -- CUHKSZ \citep{yang-nakamura-2026-cuhksz}, CPII-HK \citep{xing-etal-2026-test}, MLLP-VRAIN \citep{iranzo-sanchez-etal-2026-mllp}, PINCH-AST \citep{bentes-safka-2026-pinch}, NeMo \citep{grigoryan-etal-2026-nemo}, CUNI-POCKET \citep{sharipov-ortega-machacek-2026-pocket}, CUNI-ALIGN \citep{fuxa-machacek-2026-alignatt4llm}, and UW \citep{pong-2026-towards} -- and three proprietary API-based systems: \texttt{gemini-3.5-live-translate-preview} \citep{googledeepmind2026gemini35audio}, \texttt{gpt-realtime-translate} \citep{openai2026gptrealtimetranslate}, and \texttt{s2s-translate} \citep{gradium_s2s_translate_2026} (see Appendix~\ref{appendix_F} for an overview). All evaluations were done on five test sets: IWSLT 2025\footnote{https://iwslt.org/2025/simultaneous}, ACL, Bloomberg, Yodas, and MCIF \citep{adelani-etal-2026-speech}. See Appendix~\ref{appendix:eval_data} for a summary of the evaluation test sets. The main results are presented in Tables \ref{tab:results_de}, \ref{tab:results_ja} and \ref{tab:results_ru} (see also per-language radar charts in Appendix~\ref{appendix_E}, Figures~\ref{fig:radar_de}--\ref{fig:radar_ru}). For Hikari, Gemini-3.5-Live, GPT-Realtime and translate-s2s, all performance metrics were obtained with the help of SimulEval\footnote{https://github.com/facebookresearch/SimulEval} \citep{ma-etal-2020-simuleval}; those of the published IWSLT baselines are reproduced directly from the campaign's summary report \citep{adelani-etal-2026-speech}.


\begin{table}[ht]
\centering
\scriptsize
\setlength{\tabcolsep}{4pt}
\caption{Translation results (single-run) for English$\to$German. Systems are ranked by BLEU. StreamLAAL is reported as compute-unaware value with the compute-aware value in brackets (if available). \textbf{Hikari} (best SFT checkpoint) is shown in bold.}
\label{tab:results_de}
\begin{tabular}{@{}lrrrr@{}}
\toprule
System & BLEU & chrF & COMET & StreamLAAL \\
\midrule
\multicolumn{5}{@{}l}{\textit{IWSLT (2025)}}\\
Gemini 3.5 Live & $43.7$ & $75.0$ & $0.88$ & $2.0\;[2.0]$ \\
\textbf{Hikari} & $\mathbf{39.1}$ & $\mathbf{71.6}$ & $\mathbf{0.80}$ & $\mathbf{1.6\;[1.8]}$ \\
GPT-Realtime & $38.2$ & $70.4$ & $0.90$ & $3.9\;[3.9]$ \\
S2S-Translate & $28.0$ & $65.9$ & $0.76$ & $4.9\;[4.9]$ \\
\midrule
\multicolumn{5}{@{}l}{\textit{ACL Test Set}}\\
NEMO & $45.0$ & $71.4$ & $0.93$ & $4.4$ \\
MLLP-VRAIN & $44.9$ & $71.7$ & $0.92$ & $4.4\;[4.8]$ \\
GPT-Realtime & $36.3$ & $69.1$ & $0.90$ & $4.5\;[4.5]$ \\
CUHKSZ & $35.4$ & $64.8$ & $0.86$ & $2.5$ \\
PINCH-AST & $34.2$ & $65.4$ & $0.87$ & $5.8$ \\
CUNI-ALIGN & $34.1$ & $67.5$ & $0.87$ & $3.0\;[3.3]$ \\
\textbf{Hikari} & $\mathbf{29.1}$ & $\mathbf{65.8}$ & $\mathbf{0.79}$ & $\mathbf{1.4\;[1.6]}$ \\
Gemini 3.5 Live & $26.1$ & $53.7$ & $0.84$ & $2.9\;[2.9]$ \\
UW & $16.3$ & $47.1$ & $0.56$ & $6.6\;[6.9]$ \\
S2S-Translate & $14.5$ & $44.1$ & $0.71$ & $4.4\;[4.4]$ \\
\midrule
\multicolumn{5}{@{}l}{\textit{Bloomberg Test Set}}\\
Gemini 3.5 Live & $40.6$ & $69.8$ & $0.84$ & $4.2\;[4.2]$ \\
NEMO & $39.0$ & $66.4$ & $0.87$ & $2.2$ \\
MLLP-VRAIN & $38.9$ & $66.0$ & $0.83$ & $4.5\;[4.9]$ \\
GPT-Realtime & $35.6$ & $65.1$ & $0.91$ & $3.7\;[3.7]$ \\
\textbf{Hikari} & $\mathbf{33.6}$ & $\mathbf{65.8}$ & $\mathbf{0.80}$ & $\mathbf{2.1\;[2.4]}$ \\
CUNI-ALIGN & $29.0$ & $61.5$ & $0.74$ & $1.6\;[1.8]$ \\
PINCH-AST & $28.5$ & $59.1$ & $0.76$ & $4.7$ \\
S2S-Translate & $27.8$ & $61.7$ & $0.76$ & $4.9\;[4.9]$ \\
UW & $16.1$ & $46.5$ & $0.40$ & $5.6\;[5.9]$ \\
\midrule
\multicolumn{5}{@{}l}{\textit{Yodas Test Set}}\\
Gemini 3.5 Live & $36.8$ & $68.7$ & $0.84$ & $2.4\;[2.4]$ \\
GPT-Realtime & $36.1$ & $67.2$ & $0.87$ & $5.2\;[5.2]$ \\
\textbf{Hikari} & $\mathbf{34.1}$ & $\mathbf{64.2}$ & $\mathbf{0.83}$ & $\mathbf{2.0\;[2.2]}$ \\
NEMO & $29.6$ & $56.7$ & $0.79$ & $6.1$ \\
MLLP-VRAIN & $27.3$ & $56.8$ & $0.63$ & $4.5$ \\
CUNI-ALIGN & $27.2$ & $54.4$ & $0.72$ & $3.2\;[3.4]$ \\
S2S-Translate & $25.8$ & $61.1$ & $0.77$ & $3.2\;[3.3]$ \\
PINCH-AST & $24.2$ & $52.0$ & $0.75$ & $7.3$ \\
UW & $9.2$ & $34.6$ & $0.52$ & $8.7\;[9.0]$ \\
\midrule
\multicolumn{5}{@{}l}{\textit{MCIF Development Set}}\\
Gemini 3.5 Live & $41.1$ & $73.1$ & $0.86$ & $2.0\;[2.0]$ \\
NEMO & $37.9$ & $67.2$ & $0.93$ & $1.9$ \\
GPT-Realtime & $37.8$ & $70.1$ & $0.89$ & $4.1\;[4.1]$ \\
MLLP-VRAIN & $36.8$ & $66.5$ & $0.93$ & $3.9$ \\
\textbf{Hikari} & $\mathbf{33.5}$ & $\mathbf{67.8}$ & $\mathbf{0.80}$ & $\mathbf{1.9\;[2.0]}$ \\
CUHKSZ & $30.5$ & $60.9$ & $0.87$ & $2.0$ \\
PINCH-AST & $29.4$ & $61.8$ & $0.87$ & $5.2$ \\
S2S-Translate & $29.0$ & $63.5$ & $0.70$ & $5.0\;[5.0]$ \\
UW & $14.8$ & $44.0$ & $0.60$ & $6.7$ \\
\bottomrule
\end{tabular}
\end{table}


\begin{table}[ht]
\centering
\scriptsize
\setlength{\tabcolsep}{4pt}
\caption{Translation results for English$\to$Japanese.}
\label{tab:results_ja}
\begin{tabular}{@{}lrrrr@{}}
\toprule
System & BLEU & chrF & COMET & StreamLAAL \\
\midrule
\multicolumn{5}{@{}l}{\textit{IWSLT (2025)}}\\
\textbf{Hikari} & $\mathbf{44.6}$ & $\mathbf{49.9}$ & $\mathbf{0.66}$ & $\mathbf{3.1\;[3.1]}$ \\
Gemini 3.5 Live & $40.4$ & $48.0$ & $0.70$ & $2.0\;[2.0]$ \\
GPT-Realtime & $31.5$ & $40.4$ & $0.76$ & $4.5\;[4.5]$ \\
\midrule
\multicolumn{5}{@{}l}{\textit{ACL Test Set}}\\
Gemini 3.5 Live & $38.8$ & $46.9$ & $0.73$ & $2.4\;[2.4]$ \\
\textbf{Hikari} & $\mathbf{34.9}$ & $\mathbf{41.8}$ & $\mathbf{0.65}$ & $\mathbf{2.5\;[2.5]}$ \\
GPT-Realtime & $31.5$ & $40.9$ & $0.79$ & $5.4\;[5.4]$ \\
\midrule
\multicolumn{5}{@{}l}{\textit{Bloomberg Test Set}}\\
Gemini 3.5 Live & $31.4$ & $39.9$ & $0.64$ & $2.2\;[2.2]$ \\
\textbf{Hikari} & $\mathbf{24.7}$ & $\mathbf{35.3}$ & $\mathbf{0.50}$ & $\mathbf{2.9\;[3.1]}$ \\
GPT-Realtime & $21.6$ & $33.2$ & $0.68$ & $4.9\;[4.9]$ \\
\midrule
\multicolumn{5}{@{}l}{\textit{Yodas Test Set}}\\
\textbf{Hikari} & $\mathbf{31.6}$ & $\mathbf{40.8}$ & $\mathbf{0.73}$ & $\mathbf{2.4\;[2.6]}$ \\
Gemini 3.5 Live & $30.0$ & $39.6$ & $0.72$ & $2.0\;[2.0]$ \\
GPT-Realtime & $22.7$ & $33.2$ & $0.75$ & $8.7\;[8.7]$ \\
\midrule
\multicolumn{5}{@{}l}{\textit{MCIF Development Set}}\\
Gemini 3.5 Live & $38.7$ & $45.1$ & $0.71$ & $2.0\;[2.0]$ \\
\textbf{Hikari} & $\mathbf{34.0}$ & $\mathbf{40.9}$ & $\mathbf{0.63}$ & $\mathbf{2.4\;[2.5]}$ \\
GPT-Realtime & $29.2$ & $38.3$ & $0.77$ & $4.5\;[4.5]$ \\
\bottomrule
\end{tabular}
\end{table}


\begin{table}[ht]
\centering
\scriptsize
\setlength{\tabcolsep}{4pt}
\caption{Translation results for English$\to$Russian.}
\label{tab:results_ru}
\begin{tabular}{@{}lrrrr@{}}
\toprule
System & BLEU & chrF & COMET & StreamLAAL \\
\midrule
\multicolumn{5}{@{}l}{\textit{IWSLT (2025)}}\\
\textbf{Hikari} & $\mathbf{35.5}$ & $\mathbf{67.7}$ & $\mathbf{0.69}$ & $\mathbf{1.7\;[1.9]}$ \\
GPT-Realtime & $34.8$ & $64.0$ & $0.81$ & $4.4\;[4.4]$ \\
Gemini 3.5 Live & $31.5$ & $58.0$ & $0.78$ & $2.7\;[2.8]$ \\
\midrule
\multicolumn{5}{@{}l}{\textit{ACL Test Set}}\\
Gemini 3.5 Live & $33.4$ & $61.7$ & $0.79$ & $2.7\;[2.7]$ \\
GPT-Realtime & $28.9$ & $60.8$ & $0.80$ & $4.7\;[4.7]$ \\
\textbf{Hikari} & $\mathbf{26.6}$ & $\mathbf{61.5}$ & $\mathbf{0.70}$ & $\mathbf{1.5\;[1.7]}$ \\
\midrule
\multicolumn{5}{@{}l}{\textit{Bloomberg Test Set}}\\
Gemini 3.5 Live & $41.6$ & $68.1$ & $0.84$ & $2.7\;[2.7]$ \\
GPT-Realtime & $31.5$ & $61.5$ & $0.86$ & $5.5\;[5.5]$ \\
\textbf{Hikari} & $\mathbf{28.2}$ & $\mathbf{60.2}$ & $\mathbf{0.75}$ & $\mathbf{1.2\;[1.4]}$ \\
\midrule
\multicolumn{5}{@{}l}{\textit{Yodas Test Set}}\\
Gemini 3.5 Live & $33.2$ & $63.2$ & $0.80$ & $2.1\;[2.1]$ \\
GPT-Realtime & $29.1$ & $59.1$ & $0.79$ & $5.0\;[5.0]$ \\
\textbf{Hikari} & $\mathbf{27.8}$ & $\mathbf{57.9}$ & $\mathbf{0.77}$ & $\mathbf{1.8\;[2.1]}$ \\
\midrule
\multicolumn{5}{@{}l}{\textit{MCIF Development Set}}\\
Gemini 3.5 Live & $39.4$ & $69.1$ & $0.78$ & $2.0\;[2.0]$ \\
GPT-Realtime & $32.4$ & $62.2$ & $0.81$ & $4.4\;[4.4]$ \\
\textbf{Hikari} & $\mathbf{29.6}$ & $\mathbf{63.1}$ & $\mathbf{0.71}$ & $\mathbf{1.8\;[2.0]}$ \\
\bottomrule
\end{tabular}
\end{table}

We report BLEU \citep{papineni-etal-2002-bleu}, COMET \citep{guerreiro-etal-2024-xcomet}, and chrF \citep{popovic-2015-chrf} for quality; for latency, we use StreamLAAL \citep{papi-etal-2024-streamatt} and its compute-aware (CA) variant, which additionally accounts for hardware delay. For the API-based systems, StreamLAAL is measured end-to-end and unavoidably
includes network \emph{and} server latency; we show in Appendix~\ref{appendix:api_latency}
that this contribution is below roughly 1 s and does not affect the comparison.

\subsection{Quality-latency tradeoff}

\begin{figure}[ht!]
\vskip -0.15in
   \centering
   \includegraphics[width=1.1\columnwidth] {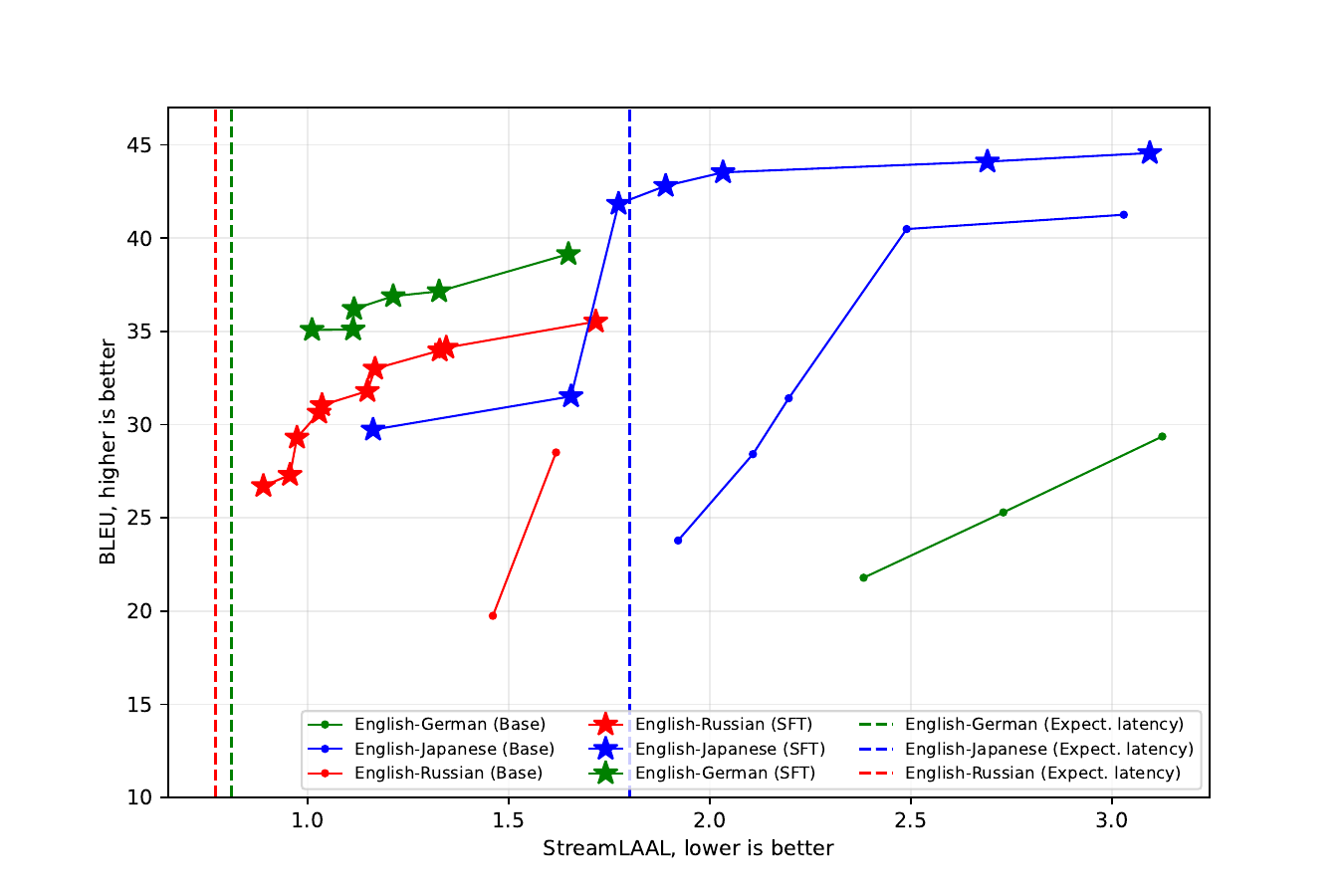}
   \caption{\label{figure2} Quality-latency tradeoff. Note how SFT shifts the Pareto frontier towards smaller latencies. Expected latencies are obtained by forcing the model to output the reference ("oracle") tokens.}
\vskip -0.1in
\end{figure}

Although Hikari does not rely on explicit policies to control this behavior, we can probe its latency response through inference-time logit biasing. By applying varying penalties to the \texttt{WAIT} token, we can modulate the model's threshold for emitting text, forcing it to generate translation earlier or allowing it to buffer more audio. As evidenced by the shift in the quality-latency curves (Figure \ref{figure2}), the SFT model achieves an overall better trade-off profile, maintaining comparable BLEU scores with latencies generally lower than the base model.

\subsection{Streaming ASR performance}

While the primary design objective of Hikari is simultaneous translation, we evaluate it as a streaming ASR system to assess the ceiling of this fully causal architecture when ASR is only an auxiliary task. Table \ref{tab:model_performance} compares our model against widely adopted streaming baselines, including Whisper-large-v3, Parakeet-0.6B \citep{pmlr-v202-xu23g}, and the recent DSM-ASR \citep{zeghidour2025streamingsequencetosequencelearningdelayed} on 4 benchmarks: TED-LIUM \citep{hernandez2018}, Meanwhile \citep{pmlr-v202-radford23a}, Earnings21 \citep{Del_Rio_2021} and Rev16 \citep{pmlr-v202-radford23a}.

\begin{table}[H]
\vskip -0.1in
\centering
\caption{Streaming ASR performance (WER \%). Results for whisper-medium.en, whisper-large-v3, Parakeet and DSM-ASR are reproduced from Table 2 of \citet{zeghidour2025streamingsequencetosequencelearningdelayed}.}
\label{tab:model_performance}
\tiny
\setlength{\tabcolsep}{4pt}
\begin{tabular}{lcccc}
\toprule
Model & TED-LIUM $\downarrow$ & Rev16 $\downarrow$ & Meanwhile $\downarrow$ & Earnings21 $\downarrow$ \\
\midrule
whisper-medium.en    & 3.9 & 6.7 & 13.0 & 12.5 \\
whisper-large-v3      & 8.1 & 11.4 & 6.1 & 11.4 \\
Parakeet-0.6B-TDT-v2  & 3.7 & 11.0 & 5.0 & 11.3 \\
DSM-ASR               & 2.9 & 12.3 & 5.7 & 10.6\\
Hikari            & 3.4 & 7.4 & 9.7 & 13.1 \\
\bottomrule
\end{tabular}
\end{table}

It is important to contextualize these results within the training constraints. State-of-the-art ASR systems are typically trained on massive datasets (e.g., 680k hours for Whisper, and over 1M hours for DSM-ASR); in contrast, Hikari was trained on approximately 62k hours -- more than an order of magnitude less data -- with ASR treated only as an auxiliary task. Despite this data disparity, Hikari still performs competitively, approaching dedicated streaming ASR systems.


\section{Ablations and other experiments}

\subsection{Effect of SFT}

\begin{table}[ht!]
\vskip 0.1in
\centering
\small
\setlength{\tabcolsep}{6pt}
\caption{Average effect of SFT across evaluation sets.
         $\Delta$BLEU$\uparrow$ and $\Delta$StreamLAAL$\downarrow$ (seconds)
         report the mean change relative to the base model;
         positive $\Delta$BLEU and negative $\Delta$StreamLAAL indicate improvement.}
\label{tab:sft_ablation}
\begin{tabular}{lrr}
\toprule
\textbf{Direction} & $\boldsymbol{\Delta}$\textbf{BLEU} $\uparrow$ & $\boldsymbol{\Delta}$\textbf{StreamLAAL (s)} $\downarrow$ \\
\midrule
en$\to$ja & $+6.65$  & $-0.31$ \\
en$\to$ru & $+8.75$  & $-0.29$ \\
en$\to$de & $+11.11$ & $-1.51$ \\
\bottomrule
\end{tabular}
\vskip -0.2in
\end{table}

The base model exhibits systematic latency drift, consistently exceeding the theoretical latency floor (Figure~\ref{figure2}, dashed lines). This occurs when extended \texttt{WAIT} sequences during silence or ambiguity push the decoder out of distribution, causing it to stall or fail to resynchronize with the incoming audio.

3While penalizing the \texttt{WAIT} token logits at inference can force lower latency, this naive approach often degrades translation quality. In contrast, our SFT strategy (Section \ref{sec:sft}) explicitly trains the model to recover from induced delays, and effectively shifts the quality-latency Pareto frontier to the left (compare thick vs. thin curves in Figure \ref{figure2}), enabling the model to "catch up" with no loss of quality. Table \ref{tab:sft_ablation} quantifies this impact: the latency is reduced (most markedly for en$\to$de), while also improving translation quality across all three language pairs (+6.65, +8.75, and +11.11 BLEU for en$\to$ja, en$\to$ru, and en$\to$de respectively). 

Figure \ref{fig:hero} shows a concrete example of how Hikari recovers from delays. Note how the final burst (from 9.2s onwards) completes a sequence of 10+ Japanese tokens in under a second. This demonstrates that the SFT "catch-up" training enables the model to reduce accumulated lag without sacrificing output quality.


\subsection{Multitask learning}

We found that training a variant of Hikari exclusively on translation objectives resulted in significant degradation in both BLEU scores and latency across all language pairs (see Table \ref{tab:asr_task_ablation}, Appendix \ref{appendix_C}), suggesting that the ASR objective provides additional acoustic grounding beneficial for low-latency simultaneous translation.

\section{Conclusions}

We presented Hikari, a policy-free, end-to-end model for simultaneous speech-to-text translation and streaming transcription. By encoding READ/WRITE decisions directly into the token distribution via a \texttt{WAIT} token mechanism, and by introducing Decoder Time Dilation to address the overrepresentation of filler tokens in training, Hikari avoids external segmentation modules entirely. Our supervised fine-tuning strategy further shifts the quality-latency Pareto frontier, yielding gains of $+6.65$--$+11.11$ BLEU alongside a reduction in StreamLAAL across all three language pairs. Despite its modest size (0.76B parameters — up to ~38× smaller than the compared IWSLT systems), Hikari attains competitive translation quality at the lowest latency among all systems on en→de and en→ru, and competitive latency on en→ja.

Several directions remain open. \textbf{Scaling.} RTF analysis (Appendix~\ref{appendix:rtf}) shows that even a Whisper-large-scale variant would remain real-time; scaling Hikari beyond the medium configuration is a natural next step. \textbf{Speech-to-speech.} Hikari's streaming decoder is a natural backbone for speech-to-speech translation via a vocoder head, in the spirit of \citet{zeghidour2025streamingsequencetosequencelearningdelayed}. \textbf{Data diversity.} Training is dominated by podcasts; broader domain coverage -- including noisy, multi-speaker, and scripted speech -- would likely close the remaining gap on out-of-distribution benchmarks.


\section*{Limitations}
\label{sec:limitations}

\paragraph{Streaming ASR.} Hikari's WER on noisy or out-of-domain benchmarks (Table~\ref{tab:model_performance}) lags behind dedicated streaming ASR systems trained on an order of magnitude more data. ASR is an auxiliary task here; the model should not be treated as a competitive standalone transcription system.

\paragraph{Fixed decoder time dilation.} $D=4$ was chosen via a coarse sweep on one language pair. A single fixed $D$ serves all speech rates uniformly; periods of unusually fast or slow speech may be handled suboptimally. An adaptive dilation mechanism is a possible extension.

\paragraph{Coverage and metrics.} Tables \ref{tab:results_de}, \ref{tab:results_ja} and \ref{tab:results_ru} show a consistent pattern: Hikari is competitive or leading on the surface metrics (BLEU, chrF) but trailing the frontier APIs and the largest IWSLT systems on COMET. We see two likely contributors. First, BLEU and chrF reward n-gram overlap with the reference, whereas COMET is sensitive to fluency and adequacy — precisely where a 0.76B model is at a disadvantage against systems 30–40× larger and against frontier APIs. Second, our targets are LLM-generated translations aligned causally at the word level, and this may bias Hikari toward literal output that scores well on surface metrics but reads less fluently. Thus, we do not interpret our BLEU results as evidence of overall superiority. Human evaluation would be best suited to adjudicate the BLEU/chrF vs COMET divergence noted in Section \ref{sec:transl_quality_and_latency}.

\paragraph{Scale.} Pre-training required 96 H100 GPUs (Section~\ref{sec:training}), placing full reproduction out of reach for many groups; we release model weights and inference code to mitigate this.

\bibliography{REFERENCES}


\clearpage
\appendix

\section{Causal Alignment and WAIT Token Interleaving}
\label{appendix_D}

\begin{algorithm}[H]
  \caption{Causal Alignment and \texttt{WAIT} Token Insertion}
  \label{alg:causal_alignment}
  \begin{algorithmic}
    \STATE {\bfseries Input:} Source word timestamps $X = \{(x_i, s_i, e_i)\}_{i=1}^M$, Target word sequence $Y = \{y_j\}_{j=1}^N$, Alignment $f$, Decoder time dilation $D$, Decoder context length $l = 1500 / D$, Token duration (ms) $\Delta t = 30000 / l$
    \STATE {\bfseries Output:} Augmented target labels $L \in \{0, \dots, V\}^l$

    \STATE Initialize $L \leftarrow [\omega, \omega, \dots, \omega]$ \COMMENT{Initialize with \texttt{WAIT} tokens}
    \STATE $P \leftarrow \text{get\_prompt\_tokens}(\text{lang, task})$ \COMMENT{Prepend task/language markers}
    \FOR{$k=1$ {\bfseries to} $|P|$}
        \STATE $L[k] \leftarrow P[k]$
    \ENDFOR

    \STATE $t_{last} \leftarrow |P|$
    \FOR{$j=1$ {\bfseries to} $N$}
        \IF{$f(j) = i \neq \emptyset$}
            \STATE \COMMENT{Aligned: Calculate index based on audio timestamp}
            \STATE $\delta \sim \mathcal{U}(0, 200)$ \COMMENT{Sample stochastic delay}
            \STATE $t_{target} \leftarrow \lceil (e_i + \delta) / \Delta t \rceil$
        \ELSE
            \STATE \COMMENT{Unaligned: Defer to the next available logical time}
            \STATE $t_{target} \leftarrow t_{last} + 1$
        \ENDIF

        \STATE \COMMENT{Ensure monotonic emission and handle potential collisions}
        \STATE $t_{start} \leftarrow \max(t_{target}, t_{last} + 1)$

        \STATE $\text{tokens} \leftarrow \text{tokenize}(y_j)$
        \FOR{$\tau \in \text{tokens}$}
            \IF{$t_{start} < l$}
                \STATE $L[t_{start}] \leftarrow \tau$
                \STATE $t_{last} \leftarrow t_{start}$
                \STATE $t_{start} \leftarrow t_{start} + 1$
            \ENDIF
        \ENDFOR
    \ENDFOR
    \STATE {\bfseries return} $L$
  \end{algorithmic}
\end{algorithm}

\section{Per-Language Radar Charts}
\label{appendix_E}

\begin{figure}[ht]
\centering
\includegraphics[width=\columnwidth]{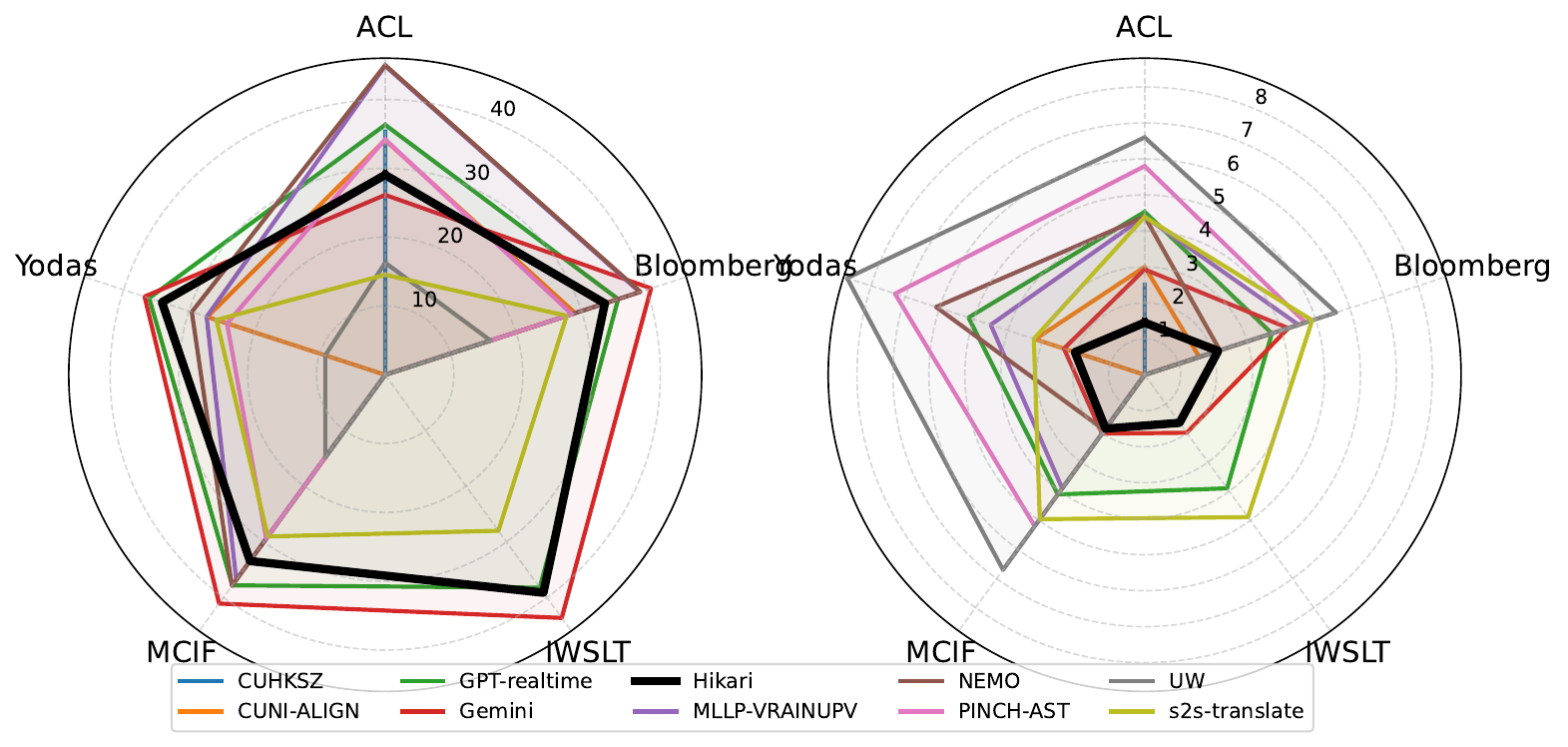}
\caption{English$\to$German: BLEU (left, higher is better) and StreamLAAL in
seconds (right, lower is better) across test sets. \textbf{Hikari} is shown
with a bold black outline.}
\label{fig:radar_de}
\end{figure}

\begin{figure}[ht]
\centering
\includegraphics[width=\columnwidth]{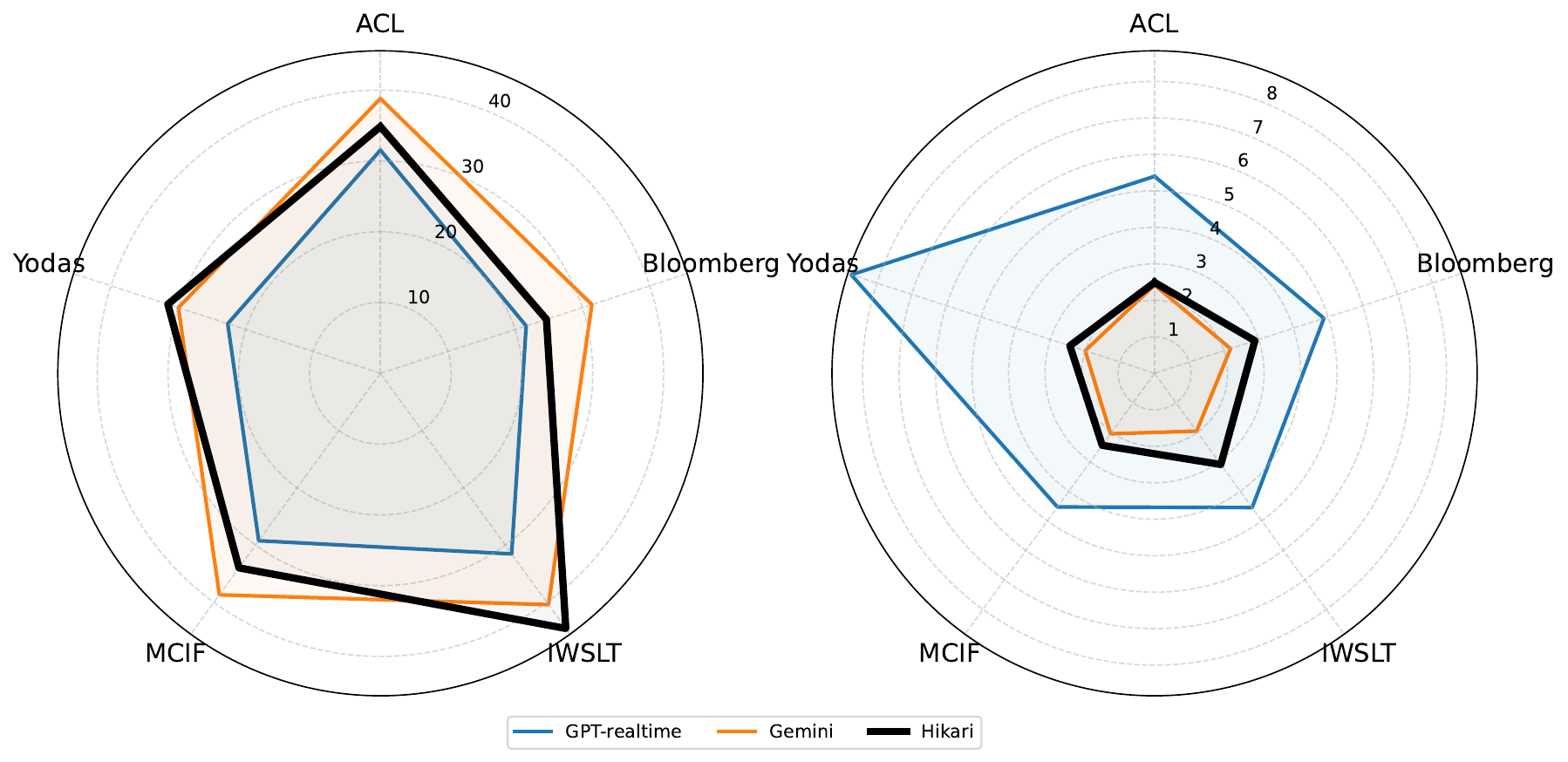}
\caption{English$\to$Japanese: BLEU (left) and StreamLAAL in seconds (right).}
\label{fig:radar_ja}
\end{figure}

\begin{figure}[ht]
\centering
\includegraphics[width=\columnwidth]{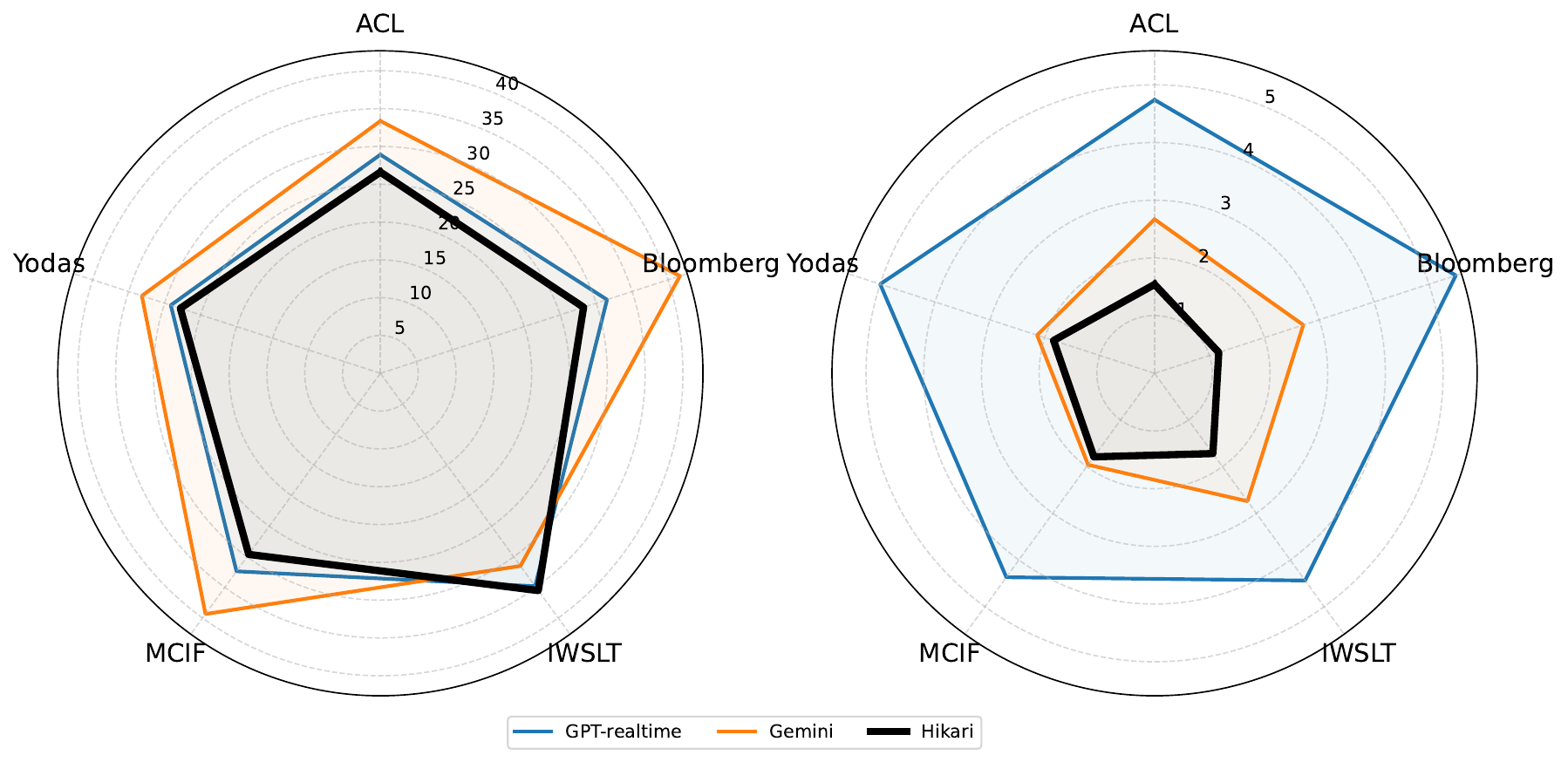}
\caption{English$\to$Russian: BLEU (left) and StreamLAAL in seconds (right).}
\label{fig:radar_ru}
\end{figure}

Figures~\ref{fig:radar_de}--\ref{fig:radar_ru} visualise BLEU and StreamLAAL
across the five evaluation test sets for each target language.
Each axis corresponds to one test set; a larger shaded area indicates higher
BLEU, while a smaller shaded area indicates lower (better) StreamLAAL.
\textbf{Hikari} is highlighted with a bold black outline.

\section{Translation Prompt}
\label{appendix_A}

\begin{tcolorbox}[title=if PREVIOUS\_SENTENCES is not None]
\small
Here's a text snippet:
\newline
\newline
\{PREVIOUS\_SENTENCES\}
\newline
\newline
Using the snippet above as context, translate another sentence into \{TGT\_LANG\}. Your translation must (1) be surrounded by the $<$TRANSLATION$>$ and $<$/TRANSLATION$>$ tags and (2) contain nothing but the translation of the source sentence. Here's the sentence for you to translate:
\newline
\newline
\{SRC\_SENTENCE\}
\end{tcolorbox}

\begin{tcolorbox}[title=if PREVIOUS\_SENTENCES is None]
\small

You have an \{SRC\_LANG\} sentence. Translate the sentence into \{TGT\_LANG\}. Your translation must be surrounded by the $<$TRANSLATION$>$ and $<$/TRANSLATION$>$ tags. Here's the sentence to translate:
\newline

\{SRC\_SENTENCE\}
\end{tcolorbox}

\section{Real-Time Factor}
\label{appendix:rtf}

\begin{figure}[ht!]
\vskip 0.1in
   \centering
   \includegraphics[width=0.99\columnwidth] {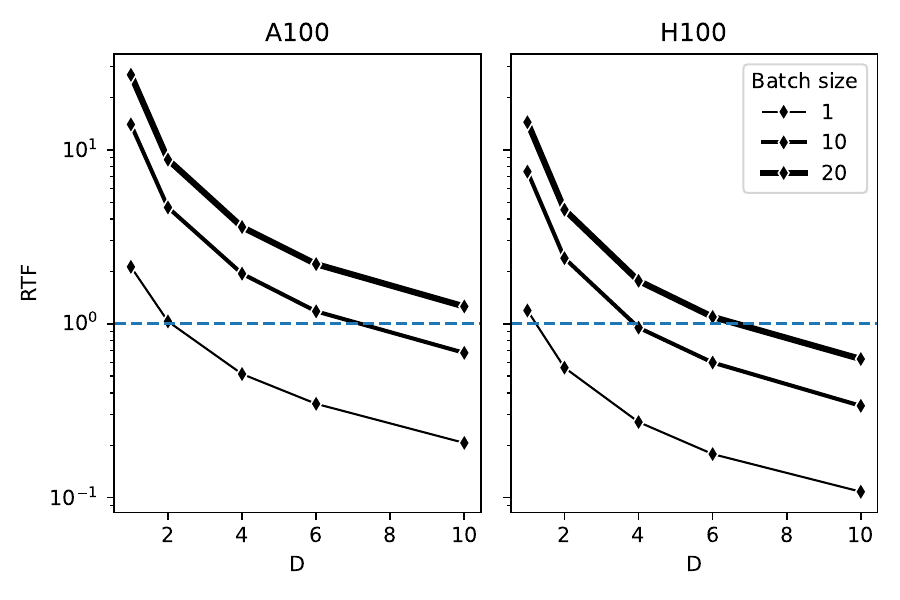}
   \caption{Real-time factor (RTF) as a function of decoder time dilation ($D$) on a single A100 or H100 GPU. The plot compares Hikari across varying batch sizes ($B \in \{1, 10, 20\}$). The dashed blue line indicates the real-time threshold (RTF $= 1.0$); values below this line indicate generation speed faster than the input audio stream.}
\vskip -0.2in
\label{fig:RTF}
\end{figure}

\section{Training and Evaluation Data}
\label{appendix:eval_data}

\begin{table}[H]
\centering
\small
\setlength{\tabcolsep}{4pt}
\caption{Training data (62.3K hours total).}
\label{tab:datasets-single}
\begin{tabularx}{\columnwidth}{l r X}
\toprule
\textbf{Name} & \textbf{Hrs} & \textbf{Description} \\
\midrule
MLS      & 13K   & 10--20 s chunks, audiobooks. \textit{CC-BY} \\
\addlinespace
CV2      & 0.2K  & $\sim$5 s phrases, noisy/accented. \textit{CC-BY} \\
\addlinespace
HQ Podcasts & 3.5K & High-quality podcasts ($\sim$1.5 h each). \textit{In-house} \\
\addlinespace
PodCrawl & 45.6K & Crawled podcasts ($\sim$1.5 h each). \textit{In-house} \\
\midrule
\textbf{Total} & \textbf{62.3K} & \\
\bottomrule
\end{tabularx}
\end{table}

\begin{table}[H]
\centering
\small
\setlength{\tabcolsep}{4pt}
\caption{Evaluation data used in this work.}
\label{tab:eval_data}
\begin{tabularx}{\columnwidth}{l r X}
\toprule
\textbf{Test set} & \textbf{Dur.} & \textbf{Description} \\
\midrule
IWSLT 2025     & $\sim$2 h   & 21 scientific oral presentations from ACL 2025 ($\sim$6 min each). Same format as the ACL set below, recorded one year earlier. \\
\addlinespace
ACL            & $\sim$2 h   & 21 scientific oral presentations from ACL 2026 ($\sim$6 min each). \\
\addlinespace
Bloomberg      & $\sim$1.5 h & Single broadcast from Asharq Business with Bloomberg; financial/business domain. \\
\addlinespace
Yodas          & $\sim$1.5 h & 5 YouTube videos (10--30 min each) from the YODAS corpus; professionally curated reference translations. \\
\addlinespace
MCIF           & $\sim$34.1 h & Development set from \citet{adelani-etal-2026-speech}. \\
\bottomrule
\end{tabularx}
\end{table}

\section{Extended Ablation Studies}
\label{appendix_B}




\subsection{The Role of ASR for Translation Performance}
\label{appendix_C}

To determine whether the ASR objective competes for model capacity or provides beneficial regularization, we conducted a task-ablation study. We trained a version of Hikari using the same dataset and hyperparameters but removed the transcription task entirely, leaving only the English-to-Japanese, German, and Russian translation objectives.

\begin{table}[ht!]
\vskip 0.2in
\centering
\small
\setlength{\tabcolsep}{6pt}
\caption{Effect of removing the ASR objective from training (averaged across evaluation sets).
         $\Delta$BLEU and $\Delta$StreamLAAL report the change relative to the full multitask model;
         negative $\Delta$BLEU and positive $\Delta$StreamLAAL indicate degradation.
         StreamLAAL CA is shown in brackets.}
\label{tab:asr_task_ablation}
\begin{tabular}{lrr}
\toprule
\textbf{Direction} & $\boldsymbol{\Delta}$\textbf{BLEU} $\uparrow$ & $\boldsymbol{\Delta}$\textbf{StreamLAAL (s)} $\downarrow$ \\
\midrule
en$\to$ja & $-7.74$ & $+4.99$ [$+4.97$] \\
en$\to$de & $-9.21$ & $+7.65$ [$+7.58$] \\
en$\to$ru & $-9.03$ & $+8.78$ [$+8.67$] \\
\bottomrule
\end{tabular}
\vskip -0.2in
\end{table}

Removing the ASR objective consistently degrades both quality and latency across all three language pairs. The BLEU drop is mildest for en$\to$ja ($-7.74$ points) and largest for en$\to$de ($-9.21$ points), while the StreamLAAL increase follows a similar but not identical pattern: smallest for en$\to$ja ($+4.99$ s) and largest for en$\to$ru ($+8.78$ s).

We hypothesize that the ASR task serves as an additional anchor for the causal encoder. By forcing the model to reconstruct the source text, it preserves acoustic features that might otherwise be lost if the model is optimized solely for semantic abstractions.



\section{Overview of Compared Systems}
\label{appendix_F}

\begin{table}[H]
\centering
\small
\setlength{\tabcolsep}{5pt}
\caption{Overview of systems compared in this work.
         Parameter counts are approximate.
         \textbf{Hikari} is our system.}
\label{tab:system_overview}
\begin{tabularx}{\columnwidth}{l c X r}
\toprule
\textbf{System} & \textbf{Proprietary} & \textbf{Code / inference released} & \textbf{Params.} \\
\midrule
CUHKSZ          & No  & No                          & 30.0 B \\
CPII-HK         & No  & No                          & 30.0 B \\
MLLP-VRAIN      & No  & No                          & 27.7 B \\
NeMo            & No  & Framework open; inference code unreleased & 27.6 B \\
PINCH-AST       & No  & No                          & 5.7 B  \\
CUNI-ALIGN      & No  & Yes                         & 5.7 B  \\
UW              & No  & No                          & 1.2 B  \\
CUNI-POCKET     & No  & Yes                         & 1.0 B  \\
\textbf{Hikari} & No  & Committed$^\dagger$         & \textbf{0.76 B} \\
\midrule
Gemini-3.5-Live & Yes & No                          & unknown \\
GPT-Realtime    & Yes & No                          & unknown \\
S2S-Translate$^*$ & Yes & No                        & unknown \\
\bottomrule
\end{tabularx}
\vspace{2pt}
{\footnotesize
$^*$~S2S-Translate is suggested by Gradium to be based on
Hibiki-Zero \citep{labiausse2026simultaneousspeechtospeechtranslationaligned};
how closely the model behind their API corresponds to the published
Hibiki-Zero is unknown.\\
$^\dagger$~Model weights and inference code will be released upon publication.
}
\end{table}

\section{Transport and server latency in the API-based systems}
\label{appendix:api_latency}

The StreamLAAL values we report for the three API systems (Gemini-3.5-Live, GPT-Realtime, S2S-Translate) include network transport and server-side computation, since these systems are reachable only as a real-time stream. This appendix shows the confound is roughly sub-second and does not change the comparison.

For locally-run systems we report a compute-unaware delay (source audio ingested at each token, reflecting only the READ/WRITE policy) and a compute-aware delay (adding GPU time). For an API we observe only the wall-clock arrival time of each token, so the two collapse to a single value — which is why every API entry in Tables \ref{tab:results_de}, \ref{tab:results_ja} and \ref{tab:results_ru} has identical brackets (e.g. GPT-Realtime 3.9 [3.9]), whereas Hikari's differ by its $\approx 0.2$ s decoder cost (1.6 [1.8]). The API figures therefore bundle transport and server computation together with the model's own waiting.

We measure the tail latency: the wall-clock gap between the last audio sample sent and the last token received. This interval cannot be attributed to the model waiting for input, so it upper-bounds transport plus server buffering. Across all test sets its median is $\approx 0.6$ s and it stays below $\approx 1$ s at the 90th percentile.

This sub-second floor is far smaller than the 2–5 s StreamLAAL of the API systems and cannot account for the multi-second gaps in Tables \ref{tab:results_de}, \ref{tab:results_ja} and \ref{tab:results_ru}. The asymmetry also favors the APIs rather than Hikari: their figures include at most $\approx 0.6$ s of non-policy latency, while Hikari's primary (compute-unaware) figure excludes its own compute. Comparing on a common compute-aware footing — Hikari's bracketed values against the API models' values — leaves the ordering unchanged. We therefore report the API models' latencies as measured and treat them as a mildly conservative estimate of each system's intrinsic latency.

\end{document}